\documentclass{isprs}
\usepackage{setspace}
\usepackage{graphicx}
\usepackage[british]{babel}
\usepackage[hang]{footmisc}
\usepackage{amsmath}
\usepackage{amssymb}
\usepackage{url}
\usepackage{multirow}

\begin{document}

\twocolumn[
\begin{center}
{\titlesize\bf Prior-Driven Enhancements in 3D Gaussian Splatting: Normals and Depths Regularization\par}
\vskip 5ex
{\normalsize Gyeonggwan Lee, Seunghwan Hong, Junghun Suh\par}
\vskip 1ex
{\normalsize
AI R\&D Team, KakaoMobility, Korea\par
\{gandan.lee, logan.sh, jude.suh\}@kakaomobility.com\par}
\end{center}
\par
\vskip 3ex
\noindent\normalsize{\bf Abstract\par\vskip \baselineskip}
{\parskip 0ex plus 2pt minus 2pt 3D Gaussian Splatting (3DGS) is a state-of-the-art technique for 3D scene rendering, offering high efficiency and excellent visual quality. However, because 3DGS relies on an initial sparse point set from Structure-from-Motion (SfM) and view-dependent properties, it can suffer from geometric inaccuracies and visual artifacts, particularly in complex scenes. To address these challenges, we propose an improved 3DGS approach that regularizes the optimization process by integrating geometric priors, including surface normals and dense depth information. Surface normal regularization improves geometric consistency by aligning Gaussian covariance with local surface structures, while dense depth priors combined with an initial points from SfM enhance per-pixel depth estimation, increasing accuracy and reducing ambiguities. These enhancements enable robust handling of diverse and complex real-world scenarios, minimizing visual distortions and improving reconstruction quality across various environments. To validate our method, we evaluate it on challenging datasets, including street-view scenes and highly reflective environments, while testing it across multiple SfM pipelines. Our results demonstrate compatibility across diverse environments and highlight the robustness of our approach. Experimental findings further show that our method enhances geometric accuracy and visual quality, establishing a reliable solution for real-time 3D scene rendering in complex environments.}
\vskip 2\baselineskip
]
\emergencystretch=2em

\section{Introduction}

3D rendering, which is the process of generating images for 3D scene from a specific point of view, is one of the fundamental research field in computer graphics, where high-quality and efficient scene synthesis is essential. It plays a key role in generating visually realistic environments while meeting real-time performance requirements. However, achieving both high-quality rendering and real-time efficiency remains a significant challenge. Thanks to the advancements in deep learning, Neural Radiance Fields (NeRF) \cite{mildenhall2021}, which represent scenes by optimizing a neural network to learn a scene's radiance field, have been introduced. More recently, 3D Gaussian Splatting (3DGS) \cite{kerbl2023} has been developed, surpassing NeRF in computational efficiency and rendering speed.

Unlike NeRF, which requires costly per-pixel inference and employs an implicit radiance field approach, 3DGS introduces a point-based representation that explicitly models a scene as a set of 3D Gaussians. By projecting these 3D Gaussians into image space and rendering the scene through a tile-based rasterization and $\alpha$ blending process, 3DGS significantly reduces computational cost and memory usage during training while enabling high-quality, real-time rendering. Due to its efficiency and effectiveness, 3DGS is rapidly gaining traction in various domains, including perception, content generation, 3D scene reconstruction and medical imaging applications (see \cite{chen2024} and references therein).

However, 3DGS also has several inherent limitations. In particular, its ability to accurately reconstruct complex real-world scenes is constrained by its reliance on an initial sparse point set from Structure-from-Motion (SfM) \cite{schonberger2016sfm} and its view-dependent appearance. In cases of insufficient or inaccurate initialization, Gaussians may be placed arbitrarily, resulting in spatial inconsistency. Furthermore, over-constructed Gaussians, which often occur in regions with view-dependent effects--- such as highly reflective surfaces--- can introduce artifacts that degrade rendering quality, leading to increased noise and instability in the reconstruction process.

To address the aforementioned challenges, we propose a method that regularizes the optimization process of 3DGS by incorporating useful geometry information obtained from an off-the-shelf monocular normal and depth estimator. We adjust the parameters of 3D Gaussians in order to align the orientation and anisotropic shape of 3D Gaussians to the underlying geometry, represented by surface normals. By leveraging surface normals, our method ensures consistent Gaussian placement in scenes featuring shiny objects and specular surfaces, thereby reducing artifacts and floater issues, leading to a more stable and accurate surface reconstruction. Moreover, dense depth priors combined with an initial points from SfM refine per-pixel depth estimation, improving geometric accuracy and resolving ambiguities caused by sparse or noisy inputs. By maintaining consistent depth cues and minimizing redundant Gaussian placements, our regularization mitigates instability in regions with repetitive patterns or low-texture surfaces.

To demonstrate the broad applicability of our enhanced 3DGS method, we validate it using diverse datasets and a wide range of SfM pipelines. Our experiments include the publicly available Tanks \& Temples dataset \cite{knapitsch2017} as well as challenging real-world data collected in complex environments. Specifically, we evaluate street-view and indoor environments captured with a Mobile Mapping System (MMS) or Ladybug6 camera system, which often exhibit repetitive patterns, strong reflections, and low-texture characteristics. These experiments confirm the effectiveness of our method across various challenging scenarios. To ensure accurate initialization and effective distribution of 3D Gaussians, we incorporate multiple SfM pipelines, handcrafted feature-based methods as COLMAP \cite{schonberger2016mvs}, deep feature-matching methods like Superpoint-Superglue \cite{detone2018,sarlin2020}, and direct feature-matching methods like LoFTR \cite{sun2021}. This facilitates faster convergence to high-fidelity representations.

Consequently, the proposed dual-prior strategy effectively addresses the limitations of existing Gaussian Splatting, as demonstrated through experiments with the aforementioned diverse datasets and SfM pipelines. This approach leads to more stable and photorealistic scene representations, making it highly suitable for real-world applications.

To summarize, our contributions are:
\begin{itemize}
\item \textbf{Prior-driven Enhancements:} We integrate surface normals and dense depth regularization into 3DGS, improving its optimization process. This enhances spatial consistency and reconstruction fidelity, resulting in more accurate and visually coherent results.
\item \textbf{Robust Real-World Validation:} The effectiveness and adaptability of our method are demonstrated through extensive empirical evaluations on diverse datasets and multiple SfM pipelines. Our approach exhibits superior performance in challenging environments including repetitive patterns, reflective surfaces, and low-texture regions.
\end{itemize}

\section{Related Works}

\subsection{3D Rendering}

The advent of NeRF marked a paradigm shift in 3D rendering by representing scenes as a continuous function parameterized by a neural network. By leveraging a Multi-Layer Perceptron (MLP) to map 3D coordinates and viewing directions to radiance and volume density, NeRF enabled high-quality novel view synthesis through differentiable volume rendering. However, its reliance on dense ray sampling and long training times posed challenges for practical applications. To address these limitations, several improvements have been proposed, including Mip-NeRF \cite{barron2021} for reducing aliasing artifacts, Instant-NGP \cite{muller2022} for accelerating training with multiresolution hash encoding, and Plenoxels \cite{fridovich2022} for achieving real-time rendering without neural networks by using a voxel-based representation. These advancements significantly enhance NeRF's efficiency by optimizing memory usage and computational resources, making it more practical for real-world applications. Despite these optimizations, NeRF remains constrained by its computationally intensive volumetric sampling and the high cost of repeatedly querying the neural network to predict color and density at multiple 3D coordinates. To overcome these limitations, alternative approaches have emerged to enhance rendering efficiency while maintaining high quality, with 3DGS standing out as a compelling solution.

3DGS mitigates NeRF's computational overhead by adopting a fundamentally different scene representation. Unlike NeRF, which relies on an MLP to model scenes as an implicit function, 3DGS represents scenes with a set of 3D Gaussians, each of which encodes spatial position, color, opacity, and an anisotropic covariance matrix. By leveraging this explicit representation, 3DGS avoids the need for dense volumetric sampling and computationally expensive neural network evaluations. Instead of NeRF's intensive rendering pipeline, 3DGS employs rasterization-based rendering, where Gaussian splats are projected onto the image plane and composited using alpha blending. This method significantly reduces computational cost and memory usage, enabling real-time rendering. Moreover, the explicit point-based representation in 3DGS allows for direct scene manipulation facilitating real-time editing and relighting, which remains a major challenge for NeRF. Recently, various follow-up studies have been introduced to enhance the efficiency and stability of 3DGS. \cite{kerbl2024} introduces a hierarchical 3D Gaussian representation that dynamically adjusts detail levels based on scene complexity, optimizing memory allocation while maintaining rendering quality. A pixel-error-driven density control method is proposed, addressing opacity handling biases and limiting unnecessary primitive generation for a more efficient and stable densification process in \cite{bulo2024}. Similar to \cite{bulo2024}, \cite{kheradmand2024} reinterprets 3DGS as a Markov Chain Monte Carlo (MCMC) process, replacing heuristic densification and pruning with a probabilistic state transition framework to enhance robustness and reconstruction accuracy. Collectively, these advancements improve memory efficiency, computational stability, and rendering performance, making 3DGS a scalable alternative to traditional NeRF models.

\subsection{Prior Regularization}

3DGS often struggles with over-reconstruction due to poor representation in sparse feature regions or highly reflective areas, leading to geometric inaccuracies and visual artifacts. Prior regularization plays a critical role in enhancing 3D geometry reconstruction and rendering fidelity by integrating structural constraints into optimization processes. Recent studies have leveraged prior information to mitigate artifacts and enhance reconstruction quality. Depth priors have been widely explored in recent works due to advances in depth estimation such as ZoeDepth \cite{bhat2023} and Metric3D \cite{yin2023}, leading to significant improvements in generalization and accuracy in \cite{chung2024,li2024a,xu2024}. FreGS \cite{zhang2024} introduced a frequency regularization method by modifying the densification strategy. To improve geometric consistency, recent studies have investigated incorporating surface normal priors into 3DGS by preserving geometry in non-textured regions \cite{li2024b}, enforcing normal consistency \cite{gao2024}, or aligning the covariance of 3D Gaussians \cite{hwang2024}.

\subsection{Point Cloud Generation for 3D Rendering}

For rendering tasks using 3DGS, well-established SfM pipelines, like COLMAP, are widely used. These pipelines ensure high-quality camera pose estimation and sparse point cloud reconstruction, forming the foundation for Gaussian initialization. SfM remains the preferred choice, particularly in applications where camera motion is available. It reconstructs 3D structures by detecting and matching feature points across multiple images to estimate both camera motion and scene geometry. Traditional hand-crafted methods, like COLMAP, offer robust feature detection and precise matching, ensuring reliable 3D reconstruction. Despite the strengths of hand-crafted SfM pipelines, their performance deteriorates in challenging scenarios, such as environments with repetitive patterns or low-texture surfaces causing errors in camera pose estimation or sparse reconstruction during the SfM phase. This degradation leads to poor input for 3DGS, resulting in incorrect geometry of the scene and introducing noise and artifacts that degrade rendering quality.

To address these limitations, deep feature-matching methods such as SuperPoint-SuperGlue (SP-SG) \cite{detone2018,sarlin2020} and direct feature-matching methods like Recurrent All-Pairs Field Transforms (RAFT) \cite{teed2020} and Local Feature Matching with Transformers (LoFTR) \cite{sun2021} have been developed, offering significant advancements over traditional feature matching techniques. SP utilizes a Convolutional Neural Network (CNN) for keypoint detection and descriptor extraction, whereas SG employs a Graph Neural Network (GNN)-based matcher to refine feature correspondences. When integrated into SfM pipelines, this combination significantly enhances camera pose estimation and sparse point cloud reconstruction, providing a more accurate foundation for applications such as Gaussian initialization in 3DGS. direct feature-matching methods, a detector-free feature matching approach, eliminate the need for explicit feature matching. RAFT utilizes a recurrent cost volume refinement approach, while LoFTR employs Transformer-based global attention mechanisms to effectively model long-range dependencies. These innovations reflect the continuous evolution of SfM techniques, presenting promising opportunities to enhance Gaussian initialization for challenging datasets, such as urban scenes with dynamic objects or natural landscapes with sparse features.

\subsection{Concurrent Work}

Recently, DN-Splatter \cite{turkulainen2025} leveraged multi-prior constraints by combining depth and normal information to address the geometric limitations of 3DGS, successfully demonstrating precise reconstructions and mesh transformations in indoor scenes. While our work also uses normal and depth constraints to mitigate 3DGS artifacts, it distinguishes itself by how priors are applied during initialization and optimization, such as in the loss term. Additionally, whereas DN-Splatter primarily focuses on indoor reconstructions with a specific set of normal-depth loss functions, our method introduces normal-depth constraints tailored for multiple SfM pipelines (COLMAP, SP-SG, LoFTR), addressing a broader range of target scenarios. As a result, our approach achieves significant improvements in visual quality and robustness, even in more challenging environments, such as outdoor street views and reflective or low-texture regions. These advancements collectively contribute to enhancing the generalizability and accuracy of 3DGS.

\section{Method}

\subsection{Preliminaries for 3D Gaussian Splatting}

3D Gaussian Splatting (3DGS) is a novel point-based differentiable approach that represents 3D geometric and radiometric scene information using a collection of 3D Gaussians. Each point is modeled as a 3D Gaussian distribution, defined as:
\begin{equation}
G(\mathbf{x};\boldsymbol{\mu},\boldsymbol{\Sigma}) = \exp\left(-\frac{1}{2}(\mathbf{x}-\boldsymbol{\mu})^{\top}\boldsymbol{\Sigma}^{-1}(\mathbf{x}-\boldsymbol{\mu})\right),
\end{equation}
where $\boldsymbol{\mu}\in\mathbb{R}^3$ denotes the Gaussian center, and $\boldsymbol{\Sigma}\in\mathbb{R}^{3\times3}$ is a positive definite covariance matrix that encodes anisotropy and orientation. To render a scene, 3DGS first projects 3D Gaussians onto image planes, transforming them into 2D Gaussians through splatting:
\begin{equation}
\boldsymbol{\mu}' = \mathbf{W}\boldsymbol{\mu}, \quad \boldsymbol{\Sigma}' = \mathbf{J}\mathbf{W}\boldsymbol{\Sigma}\mathbf{W}^{T}\mathbf{J}^{T}.
\label{eq:splat}
\end{equation}
Here, $\mathbf{W}$ represents the camera projection matrix, and the Jacobian matrix $\mathbf{J}$ models geometric distortions introduced by the transformation.

After projection, the color contributions of 2D Gaussians are composited through an $\alpha$-blending procedure. Each 2D Gaussian is associated with radiance attributes, including color $\mathbf{c}$ and opacity $o$. The final color $C(\mathbf{x})$ at a pixel $\mathbf{x}\in\mathbb{R}^2$ is computed as:
\begin{equation}
C(\mathbf{x}) = \sum_{i\in\mathcal{N}(\mathbf{x})}\alpha_i(\mathbf{x})\,\mathbf{c}_i\prod_{j<i}\bigl(1-\alpha_j(\mathbf{x})\bigr),
\end{equation}
where $\mathcal{N}(\mathbf{x})$ denotes the set of Gaussians that are projected onto the pixel $\mathbf{x}$, sorted by depth to ensure proper compositing. The per-Gaussian blending weight $\alpha_i$ is adjusted using the learned opacity $o$:
\begin{equation}
\alpha_i(\mathbf{x}) = o\exp\left(-\frac{1}{2}(\mathbf{x}-\boldsymbol{\mu}')^{\top}\boldsymbol{\Sigma}'^{-1}(\mathbf{x}-\boldsymbol{\mu}')\right),
\end{equation}
where $\boldsymbol{\mu}'\in\mathbb{R}^2$, $\boldsymbol{\Sigma}'\in\mathbb{R}^{2\times2}$ are the projected Gaussian parameters in 2D space, as defined in Eq.~(\ref{eq:splat}). These projected Gaussians are then rasterized onto the image plane, forming a set of anisotropic 2D ellipses for rendering. 3DGS is optimized using Stochastic Gradient Descent (SGD) and leverages GPU-accelerated frameworks for efficient computation.

\subsection{SfM for 3D Gaussian Splatting}

The optimization speed and final rendering quality of 3DGS are strongly influenced by the quality of the initial Gaussians derived from the 3D pointcloud and camera poses. To obtain these initial Gaussians, SfM techniques are employed to generate a sparse point cloud and accurate camera poses. For example, COLMAP, a widely used SfM solution, utilizes SIFT-based feature detection and matching. However, its performance degrades in low-texture or repetitive-pattern regions, reducing reconstruction quality. To address these challenges, SP-SG enhances robustness in feature extraction and image matching against lighting variations and image distortions using a graph neural network. Yet, it still struggle in feature-sparse areas. In contrast, LoFTR learns pixel-wise correspondences directly using Transformer-based architecture. By leveraging the global receptive field of transformers, LoFTR excels in low-texture regions where traditional feature detectors struggle. To evaluate the impact of initial Gaussians generated via SfM on 3DGS performance, this paper compares COLMAP, SP-SG, and LoFTR.

\subsection{Normal Prior Regularization}

Surface normals can be estimated using various methods, including models like Omnidata \cite{kar2022}, Metric3D \cite{yin2023}, and DSINE \cite{bae2024}, which guide the alignment of Gaussians with local surface geometry. Aligning Gaussians with surface normals improves their placement on complex surfaces, ensuring the creation of more accurate and well-aligned Gaussians that capture fine details and maintain geometric consistency. Incorporating surface normals enhances the accuracy of the geometric representation and ensures improved visual fidelity. To achieve this, two key components of Normal Prior Regularization are utilized:

\begin{enumerate}
\item \textbf{Geometry-Aware Initialization:} 3D Gaussians are initialized by aligning their orientation and scale with the predicted surface normals to reflect local surface variations. This geometry-aware setup not only accelerates convergence, but also enhances stability during optimization, providing a robust starting point for the process.
\item \textbf{Normal-Consistency Loss:} This loss function is designed to align the orientation and scale of covariance of Gaussians with the underlying 3D surface geometry. Building on \cite{hwang2024}, which shows that flattening covariance along the surface normal can reduce artifacts, we define the normal consistency loss as $\mathcal{L}_{\text{normal}} = \lambda\mathcal{L}_{\text{axis}} + (1-\lambda)\mathcal{L}_{\text{scale}}$, where $\lambda$ is a hyperparameter. The orientation loss, $\mathcal{L}_{\text{axis}}$, and the scale loss, $\mathcal{L}_{\text{scale}}$, are defined as follows:
\begin{equation}
\resizebox{0.80\linewidth}{!}{$\displaystyle\mathcal{L}_{\text{axis}} = \frac{1}{N}\sum_{i=1}^{N}\sum_{j\in\{0,1,2\}}\left|\left(\tilde{R}^{(i)}_{\mathrm{G}}[:,j]\cdot\mathbf{n}^{(i)}\right)\right|$}
\end{equation}
\begin{equation}
\resizebox{0.80\linewidth}{!}{$\displaystyle\mathcal{L}_{\text{scale}} = \frac{1}{N}\sum_{i=1}^{N}\sum_{j\in\{0,1,2\}}\tilde{s}^{i}_{G}[j]\left|\left(\tilde{R}^{(i)}_{\mathrm{G}}[:,j]\cdot\mathbf{n}^{(i)}\right)\right|$}
\end{equation}
Here, $N$ is the total number of Gaussians, and $\tilde{R}^{(i)}_{\mathrm{G}}[:,j]$ and $\tilde{s}^{(i)}_{\mathrm{G}}[:,j]$ represent the orientation and scale of the $j$th axis of covariance for $i$-th Gaussian, respectively. $\mathbf{n}^{(i)}$ is the surface normal estimated through normal prediction network related to $i$-th Gaussian.
\end{enumerate}

Through geometry-aware initialization and normal-consistency loss, normal prior regularization improves the quality of the covariance parameters of 3D Gaussians by aligning them with surface geometry. This method effectively addresses challenges posed by complex surfaces and occlusions, enhancing both geometric fidelity and visual quality.

\subsection{Depth Prior Regularization}

Depth prior regularization enhances the representation of the scene's geometric structure by utilizing dense depth maps derived from images. By capturing depth information for every pixel, this approach allows for a more precise depiction of depth compared to the sparse information typically used in conventional Gaussian Splatting.

However, there is a scale issue with the depth map, $D_{\text{dense}}$, obtained from the depth prediction network. To solve this, we use the sparse depth map, $D_{\text{sparse}}$, which is generated by projecting SfM points onto the images. By adjusting the scale of $D_{\text{dense}}$ to match $D_{\text{sparse}}$, similar to prior work \cite{chung2024}, we obtain the optimized depth map, $D_{\text{guide}}$, which serves as a regularized dense depth map.

Building on this, a depth-consistency loss is defined using L1 distance as follows: $\mathcal{L}_{\text{depth}} = \left\lVert(D_{\text{guide}} - D)\right\rVert_1$, where $D$ is the rendered depth map from Gaussian splatting.

\section{Experiments}

To evaluate the performance of our 3DGS method, we conducted experiments not only on the Tanks \& Temples dataset but also on additional datasets that exhibit challenging characteristics such as repetitive patterns, reflections, and low-texture surfaces. Furthermore, we applied and compared multiple SfM pipelines to quantitatively assess the impact of Gaussian initialization quality on the final reconstruction results. Through extensive experiments across diverse environments and conditions, we confirmed that our surface normal regularization effectively reduces geometric distortions, while our dense depth map regularization improves depth estimation accuracy.

\begin{figure}[ht!]
\begin{center}
\begin{tabular}{@{}c@{\hspace{4mm}}c@{}}
\includegraphics[height=2.9cm]{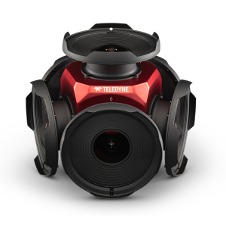} &
\includegraphics[height=2.9cm]{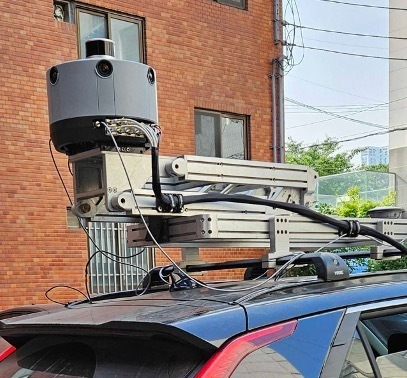} \\
\small (a) Ladybug6 & \small (b) Mobile Mapping System \\
 & \small (MMS)
\end{tabular}
\caption{Hardware for Data Collection}
\label{fig:hardware}
\end{center}
\end{figure}

\subsection{Experimental settings}

\textbf{Datasets}\quad We conduct our experiments on both publicly available datasets and custom indoor and outdoor scenes, which exhibit unique characteristics and challenges. Specifically, we use two scenes from the Tanks \& Temples dataset \cite{knapitsch2017} ---Train and Horse---which are commonly used in 3DGS research. Additionally, we use real-world data, which we captured ourselves, to demonstrate the superior performance of our method in both indoor and outdoor environments. We first collected data in underground parking lots using a Ladybug6 camera system shown in Figure~\ref{fig:hardware}a, which captures images with six cameras (see \cite{ladybug6} for more information). Among them, we utilized the three cameras positioned at the front and sides to construct the dataset. Each image has a resolution of $2992\times4096$ and a field of view (FOV) of 85.9 degrees. Images are captured at 1 FPS and post-processed to correct image distortion. Our indoor dataset presents significant challenges due to low-texture surfaces, reflections, and varying illumination. Additionally, we collected outdoor scene data by capturing street-view environments using a Mobile Mapping System (MMS) mounted on a moving vehicle.

Our MMS shown in Figure~\ref{fig:hardware}b simultaneously captures rectangular images using six cameras at fixed distance intervals. The collected images are processed to generate equirectangular images, then we extract seven rectangular images from each equirectangular image, with each having a 90$^\circ$ (deg) field of view (FOV). These extracted images are positioned at 45$^\circ$ (deg) intervals, covering all directions except the rear. The final extracted images have a resolution of $1080\times1080$. Since these street-view scenes include dynamic obstacles, such as cars and pedestrians, and repetitive patterns, such as road markings and similar building facades, feature matching and pose estimation become challenging, leading to an inaccurate and sparse point cloud.

\textbf{Evaluation metrics}\quad To compare results across multiple SfM pipelines, we use the mean reprojection error (MRE) as a metric to evaluate geometric consistency. Additionally, we assess reconstruction quality using three standard evaluation metrics: Peak Signal-to-Noise Ratio (PSNR), Structural Similarity Index Measure (SSIM), and Learned Perceptual Image Patch Similarity (LPIPS). These metrics provide a comprehensive analysis by capturing both objective accuracy and perceptual fidelity, demonstrating the improvements achieved over the baseline.

\textbf{Implementation Details}\quad In this study, we establish conventional 3DGS \cite{kerbl2023} as the baseline for comparison. We also analyze the impact of different SfM methods on initial Gaussian placement and optimization by comparing COLMAP, SP-SG, and LoFTR. For our experiments, we use their official implementations. To obtain priors for regularizing the optimization process, we utilize the pre-trained Metric3D \cite{yin2023} network for monocular depth and normal estimation. Our final optimization loss function is defined as:
\begin{equation}
\mathcal{L} = (1-\lambda_1)\mathcal{L}_{\text{color}} + \lambda_1\mathcal{L}_{\text{D-SSIM}} + \lambda_2\mathcal{L}_{\text{normal}} + \lambda_3\mathcal{L}_{\text{depth}}
\end{equation}
where the first two loss terms, $\mathcal{L}_{\text{color}}$ and $\mathcal{L}_{\text{D-SSIM}}$, correspond to the original 3DGS losses in \cite{kerbl2023}. We set $\lambda_1$, $\lambda_2$, and $\lambda_3$ as 0.2, 0.01, and 0.01 respectively.

\begin{table*}[t]
\centering
\small
\setlength{\tabcolsep}{3.3pt}
\begin{tabular}{ll|cccc|cccc|cccc}
\hline
 & & \multicolumn{4}{c|}{COLMAP} & \multicolumn{4}{c|}{SP-SG} & \multicolumn{4}{c}{LoFTR} \\
Datasets & Method & MRE$\downarrow$ & PSNR$\uparrow$ & SSIM$\uparrow$ & LPIPS$\downarrow$ & MRE$\downarrow$ & PSNR$\uparrow$ & SSIM$\uparrow$ & LPIPS$\downarrow$ & MRE$\downarrow$ & PSNR$\uparrow$ & SSIM$\uparrow$ & LPIPS$\downarrow$ \\
\hline
\multirow{2}{*}{Train} & 3DGS & \multirow{2}{*}{0.75} & 21.10 & 0.802 & 0.218 & \multirow{2}{*}{1.40} & 21.10 & 0.750 & 0.282 & \multirow{2}{*}{0.79} & 20.97 & 0.767 & 0.274 \\
 & Ours & & 21.97 & 0.799 & 0.252 & & 21.32 & 0.749 & 0.287 & & 21.11 & 0.759 & 0.291 \\
\hline
\multirow{2}{*}{Horse} & 3DGS & \multirow{2}{*}{0.71} & 24.18 & 0.889 & 0.239 & \multirow{2}{*}{1.31} & 21.01 & 0.802 & 0.239 & \multirow{2}{*}{0.80} & 23.39 & 0.870 & 0.174 \\
 & Ours & & 25.50 & 0.903 & 0.153 & & 21.12 & 0.801 & 0.246 & & 24.80 & 0.881 & 0.165 \\
\hline
\multirow{2}{*}{Parking lots} & 3DGS & \multirow{2}{*}{Invalid} & - & - & - & \multirow{2}{*}{1.37} & 28.08 & 0.828 & 0.424 & \multirow{2}{*}{0.64} & 29.33 & 0.842 & 0.410 \\
 & Ours & & - & - & - & & 28.37 & 0.829 & 0.427 & & 29.07 & 0.840 & 0.414 \\
\hline
\multirow{2}{*}{Street-view} & 3DGS & \multirow{2}{*}{Invalid} & - & - & - & \multirow{2}{*}{1.09} & 18.33 & 0.468 & 0.410 & \multirow{2}{*}{0.56} & 22.28 & 0.729 & 0.331 \\
 & Ours & & - & - & - & & 19.00 & 0.479 & 0.492 & & 22.49 & 0.722 & 0.340 \\
\hline
\end{tabular}
\caption{Quantitative comparison of 3DGS, Our Method, and SfM-Rendering Correlation - A lower Mean Reprojection Error (MRE) generally corresponds to a higher PSNR and SSIM, while its correlation with LPIPS is relatively weak. Among the SfM pipelines, COLMAP and LoFTR achieve relatively low MRE, whereas SP-SG exhibits a higher MRE. For the parking lot and street-view datasets, SfM with COLMAP failed to reconstruct the scenes, resulting in invalid outputs. The proposed method demonstrates improved PSNR and SSIM performance across the COLMAP, SP-SG, and LoFTR approaches.}
\label{tab:results}
\end{table*}

\subsection{Results}

\textbf{SfM on Gaussian Initialization and 3DGS}\quad We first compare the Mean Reprojection Error (MRE) of point clouds generated across multiple SfM pipelines, as shown in Table~\ref{tab:results}. The results indicate that a lower MRE consistently corresponds to a higher PSNR and SSIM, while its correlation with LPIPS is relatively weak. COLMAP achieves relatively low MRE on official benchmark datasets such as Train and Horse; however, it fails to generate a point cloud for parking lot and street-view datasets. These findings suggest that COLMAP performs well in feature-rich environments but struggles in datasets with wide baselines and reflective surfaces, where its performance significantly deteriorates. In contrast, both SP-SG and LoFTR successfully reconstruct point clouds for our collected datasets, demonstrating greater robustness in SfM tasks, despite producing slightly lower-quality results on the Train and Horse datasets. These experiments highlight that LoFTR, as direct feature-matching methods, could serve as a strong alternative for challenging scenarios where COLMAP struggles to perform SfM.

\textbf{Quantitative Comparison}\quad Our method consistently achieves higher PSNR than 3DGS across multiple datasets and SfM pipelines. As shown in Table~\ref{tab:results}, for the Train dataset with COLMAP, our method improves PSNR from 21.10 to 21.97, and for the Horse dataset, PSNR increases from 24.18 to 25.50. Similarly, for the Street View dataset with LoFTR, PSNR improves from 22.28 to 25.49. These results suggest that the optimized depth and normal estimation in our approach more accurately recovers image intensities, reducing numerical differences from the ground truth. However, for the parking lots dataset with LoFTR, PSNR degrades from 29.33 to 29.07. This decline is primarily attributed to light blurring, where artificial lighting reflections on surfaces such as floors introduce distortions, negatively impacting performance. In most cases, SSIM remains comparable, indicating that both methods maintain similar structural integrity in the rendered images. Since SSIM measures how well structural details, edges, and textures are preserved, the results suggest that the normal and depth estimated by Metric3D do not capture fine details in rendered scenes as effectively. Additionally, an increase in LPIPS in some cases suggests a slight degradation in perceptual quality, as higher LPIPS values indicate greater perceptual differences from the ground truth. The use of low-quality normal and depth estimations can oversmooth high-frequency details, such as grass, foliage, gravel, and clouds. Furthermore, since LPIPS considers perceptual similarity, even slight deviations in shading and reflections can increase LPIPS. This effect is exacerbated when Gaussians are too large or overly smooth, leading to the loss of small texture variations. While 3D Gaussians blend color and opacity smoothly, which is beneficial for noise reduction, this process can also lead to texture oversmoothing. Nevertheless, scene structures, edges, and object placements remain well-preserved, as evidenced by high PSNR and SSIM scores.

\begin{figure*}[t]
\begin{center}
\includegraphics[width=0.92\textwidth]{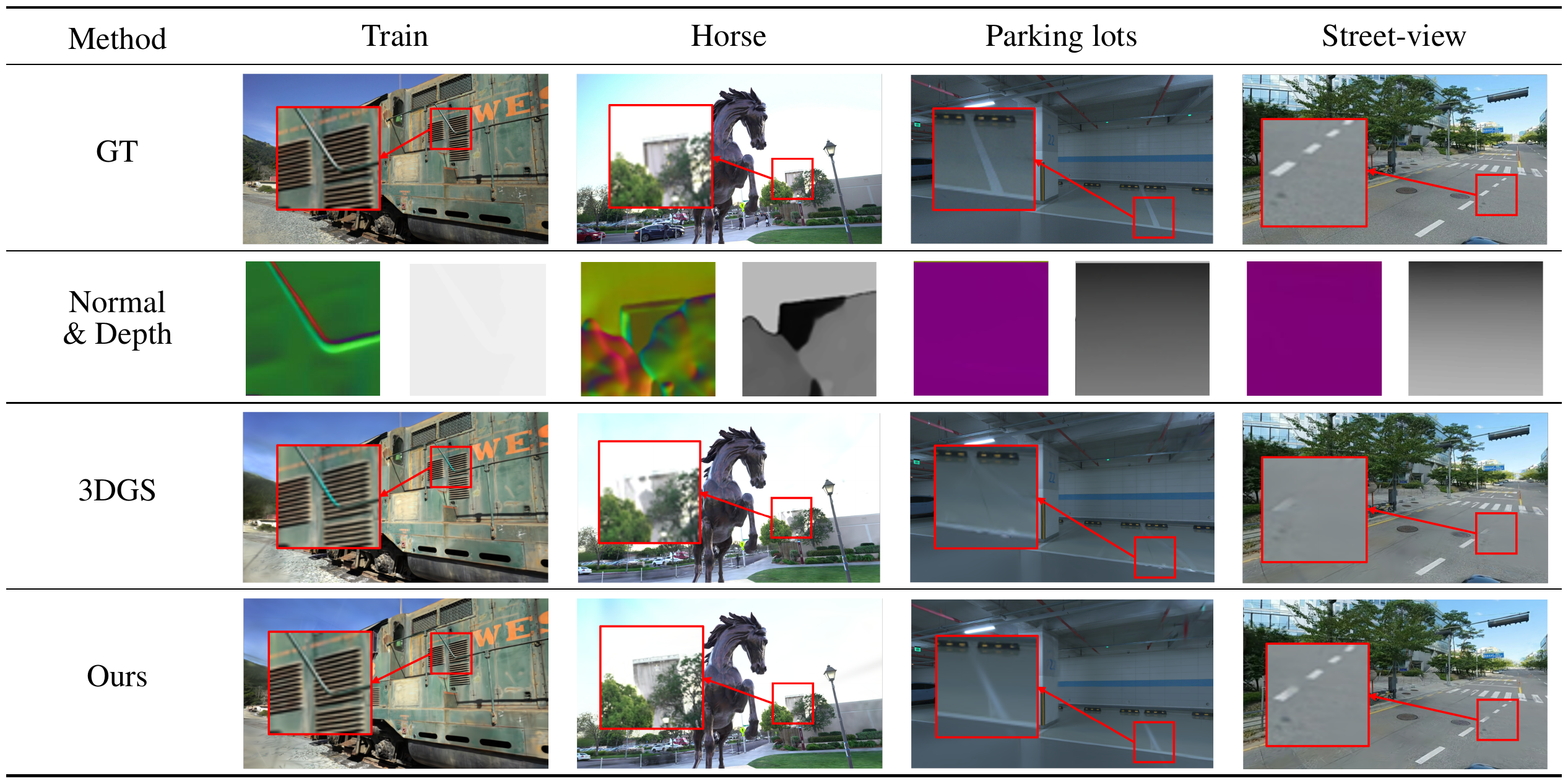}
\caption{Qualitative comparison of 3DGS and our method - Our method demonstrates improved rendering quality over 3DGS by leveraging surface normals and dense depth as prior information. This prior information enhance Gaussian alignment and depth representation. Normal \& Depth information for parking lots and street-view, flat surfaces, is represented appropriately.}
\label{fig:qualitative}
\end{center}
\end{figure*}

\textbf{Qualitative Comparison}\quad As shown in Figure~\ref{fig:qualitative}, our approach demonstrates a notable visual improvement in rendering quality compared to 3DGS. In particular, by leveraging surface normals as a prior to guide the alignment of Gaussians on flat surfaces (e.g., roads and walls), Gaussians are distributed more distinctly, leading to improved rendering accuracy. Moreover, with the well-aligned Gaussians, the road markings in the Street View dataset and the edges of walls in the Parking Lots dataset are rendered more clearly. Notably, in the Horse dataset, our approach exhibits significant improvements in the representation of distant surfaces, such as buildings located far from the camera viewpoint. These improvements suggest that the incorporation of normal and depth information significantly influences the quality of rendered scenes, leading to more accurate geometry representation and improved visual fidelity.

\section{Conclusion}

In this paper, we introduced a prior-driven enhancement approach for 3D Gaussian Splatting (3DGS) by incorporating surface normals and dense depth priors to improve geometric accuracy and visual quality. Our method regularizes the optimization process by aligning Gaussian covariance with local surface structures and refining per-pixel depth estimation. These enhancements effectively reduce artifacts, improve reconstruction consistency, and enable more robust handling of complex real-world scenes. Through extensive evaluations on challenging datasets, including highly reflective environments and street-view scenes, and across multiple SfM pipelines, our approach demonstrates superior performance in terms of geometric consistency and rendering quality. Experimental results confirm that our method enhances accuracy, stability, and adaptability, making 3DGS a more reliable solution for real-time 3D scene rendering in diverse environments.

\section{Acknowledgment}

This work was supported by the Institute of Information \& communications Technology Planning \& Evaluation (IITP) grant funded by the Korea government (MSIT) (No. RS-2023-00229833, Development of Intelligent Teleoperation Technology for Cloud-based Autonomous Vehicle Errors and Limit Situation). This work was also supported by Kakaomobility.

{
\begin{spacing}{1.17}
\normalsize
\bibliography{references}
\end{spacing}
}

\end{document}